%% file: main.tex
\documentclass[conference]{ieeeconf}  
\usepackage{amsmath,amsfonts}
\usepackage{algorithmic}
\usepackage{algorithm}

\usepackage{array}
\usepackage[caption=false,font=normalsize,labelfont=sf,textfont=sf]{subfig}
\usepackage{textcomp}
\usepackage{stfloats}
\usepackage{url}
\usepackage{verbatim}
\usepackage{graphicx}
\usepackage{multirow}
\usepackage[normalem]{ulem}

\usepackage{orcidlink}
\usepackage{hyperref}
\hypersetup{colorlinks, linkcolor = blue, citecolor=blue}
\usepackage{cite}

\usepackage{amsthm}
\usepackage{amsmath}

\DeclareMathOperator*{\argmin}{arg\,min}

\let\labelindent\relax
\usepackage{enumitem}

\usepackage{ragged2e}

\usepackage{booktabs}  
\usepackage{makecell}  
\usepackage{colortbl}  

\theoremstyle{definition}

\theoremstyle{remark}

\usepackage{subfig}

\IEEEoverridecommandlockouts                              

\begin{document}
\title{\bf
Personalized Autonomous Driving via Optimal Control 
\\with Clearance Constraints from Questionnaires
}

\author{Yongjae Lim, Dabin Kim, and H. Jin Kim
\thanks{This work was supported by the IITP(Institute of Information \& Communications Technology Planning \& Evaluation)-ITRC(Information Technology Research Center) grant funded by the Korea government(Ministry of Science and ICT)(IITP-2025-RS-2024-00437268)}
\thanks{Yongjae Lim is with the Department of Mechanical Engineering, Seoul National University, Seoul 08826, South Korea. (e-mail: yongjae.lim@snu.ac.kr)}
\thanks{Dabin Kim is with the Department of Aerospace Engineering, Seoul National University, Seoul 08826, South Korea. (e-mail: dabin404@snu.ac.kr)}
\thanks{H. Jin Kim is with the Department of Aerospace Engineering, Seoul National University, Seoul 08826, South Korea. (e-mail: hjinkim@snu.ac.kr)}
}


\maketitle
\thispagestyle{empty}
\pagestyle{empty}

\begin{abstract}
Driving without considering the preferred separation distance from surrounding vehicles may cause discomfort for users.
To address this limitation, we propose a planning framework that explicitly incorporates user preferences regarding the desired level of safe clearance from surrounding vehicles. 
We design a questionnaire purposefully tailored to capture user preferences relevant to our framework, while minimizing unnecessary questions.
Specifically, the questionnaire considers various interaction-relevant factors, including the
surrounding vehicle's size, speed, position, and maneuvers of surrounding vehicles, as well as the maneuvers of the ego vehicle. 
The response indicates the user-preferred clearance for the scenario defined by the question and is incorporated as constraints in the optimal control problem. 
However, it is impractical to account for all possible scenarios that may arise in a driving environment within a single optimal control problem, as the resulting computational complexity renders real-time implementation infeasible.
To overcome this limitation, we approximate the original problem by decomposing it into multiple subproblems, each dealing with one fixed scenario.
We then solve these subproblems in parallel and select one using the cost function from the original problem.
To validate our work, we conduct simulations using different user responses to the questionnaire.
We assess how effectively our planner reflects user preferences compared to preference-agnostic baseline planners by measuring preference alignment.
\end{abstract}
\input{Introduction}

\input{Problem_statement}

\input{Method}

\input{Result}

\input{Conclusion}

\bibliographystyle{ieeetr}
\bibliography{ref}

\end{document}

%% file: Introduction.tex
\section{Introduction}
Incorporating user preferences into the driving process is essential for achieving user-friendly driving.
In particular, user-preferred distance from surrounding vehicles is a significant factor for psychological comfort.
However, most prior works \cite{andersson2016model, kamel2017robust, tordesillas2019faster, adajania2022multi,de2024topology,zheng2023real} have overlooked this aspect, resulting in psychological discomfort and reduced trust in the autonomous driving system (ADS) from the users' perspective \cite{basu2017you}.

To incorporate user preferences into autonomous driving, the data-driven approach \cite{wang2018driving, huang2021personalized, wang2022gaussian, wen2023modeling, wang2017driving, kim2021driving, schrum2024maveric} is primarily employed.
This approach leverages user driving datasets or constructs custom ones to identify implicit features that influence driving style and integrates them into planning.
However, these datasets may not align with or fully capture the user preferences that the algorithm intends to model, making it unclear whether the preferences are reflected.

\begin{figure}[t]
\captionsetup{singlelinecheck = false, font=small, labelfont=bf, skip = 0pt}
\captionsetup[subfloat]{singlelinecheck = true, labelfont=scriptsize,textfont=scriptsize}
\centering
\vspace{5pt}
\includegraphics[height=8.0cm]{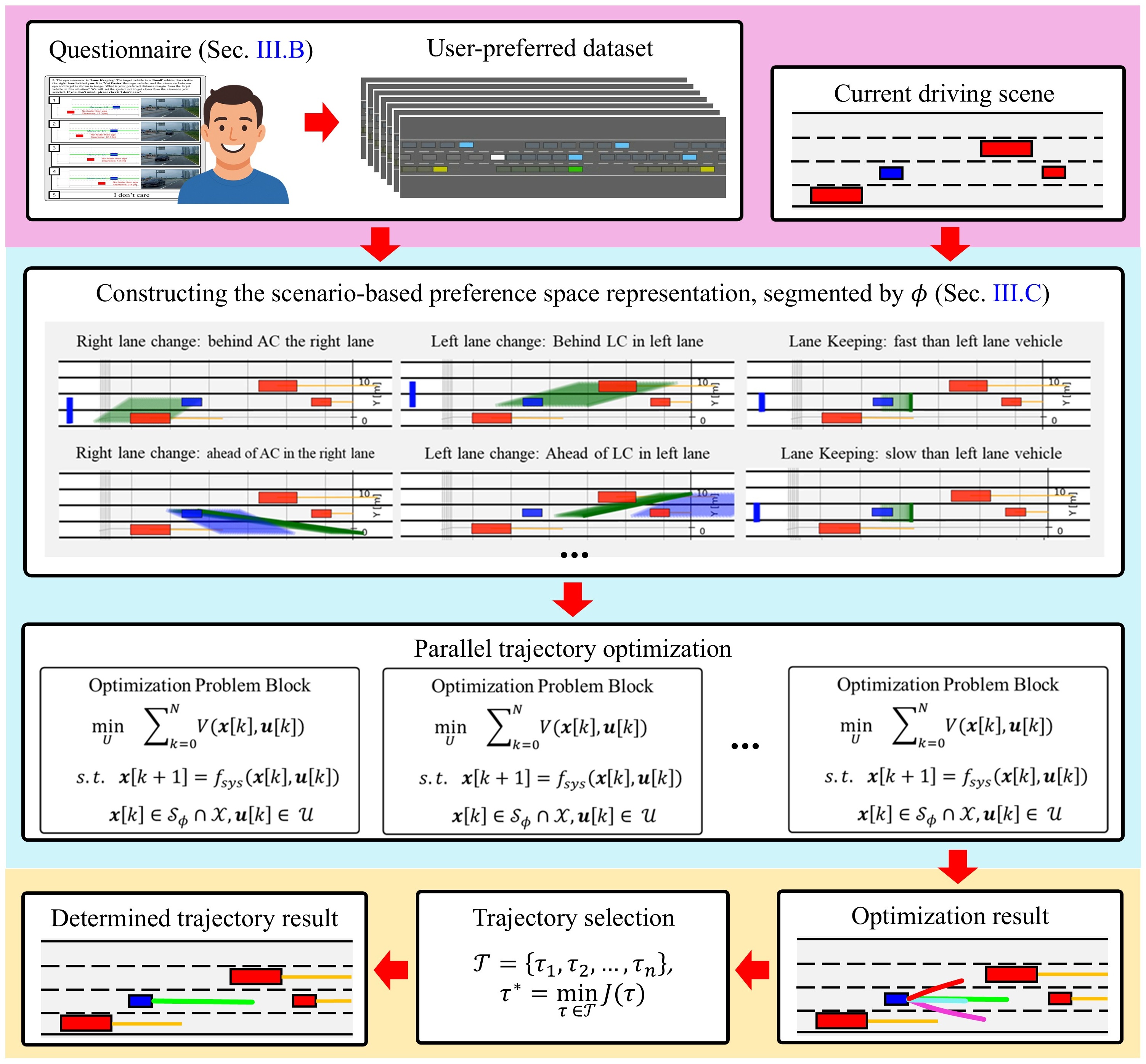}
\caption{
Our framework consists of three stages, illustrated by the \textcolor{magenta}{magenta}, \textcolor{cyan}{cyan}, and \textcolor{orange}{orange} shaded regions, respectively.
\textcolor{magenta}{The first stage (questionnaire stage)} collects the user-preferred clearance margin through a questionnaire. 
\textcolor{cyan}{The second stage (parallel trajectory optimization stage)} takes the questionnaire responses and the current driving scenario as inputs, then decomposes the original problem into scenario-specific subproblems. \textcolor{orange}{Finally, in the third stage (selection stage)}, these subproblems are solved in parallel, and one is selected using the cost function of the original problem.
}
\label{fig:framework}
\vspace{-15pt}
\end{figure}

In this study, we propose a planning framework that explicitly incorporates user-preferred clearance, defined as the desired level of safe distance from surrounding vehicles, as shown in Fig.~\ref{fig:framework}.
Unlike prior works, our study begins by explicitly defining how user preferences will be integrated into the system and designing a questionnaire based on this integration strategy.
Specifically, each question in the questionnaire considers various factors that influence user preferences in the vehicle interaction, including the surrounding vehicle's size, speed, position, and maneuver, as well as the ego vehicle's maneuver.
The question asks users about their \textit{preferred clearance margins} (i.e., the minimum acceptable longitudinal distance users wish to maintain from surrounding vehicles) within the \textit{scenario space} (i.e., the entire set of possible driving scenarios).
To address the entire scenario space with a finite number of questionnaire items, we decompose and discretize all possible scenarios into a manageable set of representative cases.
As a result, the designed questionnaire minimizes unnecessary or ambiguous questions and enables direct utilization of the responses within the planning framework.

The user responses to the questionnaire are then directly used to define a \textit{preference space} (i.e., region defined by the user-preferred clearance margin on the road) for the given driving scenario.
To reflect this space in planning, we formulate the optimal control problem (OCP) with constraints derived from the preference space. 
This explicit formulation enables direct and quantitative measurement of preferences through the evaluation of whether the vehicle state remains within the preference space.
However, achieving the objective of the formulated OCP involves diverse scenarios in the driving scene.
Methods such as Mixed Integer Programming (MIP) \cite{wolsey2007mixed} and Signal Temporal Logic (STL) \cite{karlsson2021encoding} can be used to handle these scenarios within a single OCP, but they suffer from computational burden.

To address this limitation, we approximate the original OCP by decomposing it into multiple \textit{scenario-specific subproblems}. 
This approximation allows us to handle the original OCP in real-time by considering the fixed scenarios for each subproblem.
We employ parallel trajectory optimization \cite{zhou2021raptor, de2024topology} to solve the subproblems, and then select the most appropriate trajectory among the solutions based on the cost function of the original OCP.

To validate our approach, we administer the questionnaire to generate a sample set of responses representing different driver types (e.g., aggressive or conservative). 
Based on these responses, we conduct comparisons with various baselines and perform a quantitative evaluation of preference alignment, which demonstrates that our planner successfully captures user preferences without compromising safety.

The main contributions are summarized as follows:
\begin{enumerate}
\item We propose a new planning framework that explicitly incorporates user preferences through user-preferred clearance, defined as the desired level of safe clearance from surrounding vehicles.

\item We systematically design a questionnaire that captures user preferences in the vehicle interaction across entire road driving scenarios, while remaining directly incorporable as constraints in OCP for preference-aware driving.

\item Since considering all scenarios in a single OCP is intractable, we decompose the problem into multiple subproblems, each with a fixed scenario, enabling real-time implementation.
\end{enumerate}

\section{Related works}
\subsection{Optimization-based planning in autonomous driving}
In autonomous driving, the planning module generates a safe trajectory for the vehicle to reach its destination.
Among various approaches, the optimization-based approach achieves this by formulating trajectory planning as a constrained optimization problem.
\cite{andersson2016model, kamel2017robust, tordesillas2019faster} consider collision avoidance with obstacles as constraints to generate safe trajectories.
Since there can be multiple feasible trajectories in the current situation, single-trajectory optimization may converge to a local optimum.
To address this issue, \cite{adajania2022multi, zheng2023real, de2024topology} plan multiple safe trajectories for the current situation using parallel trajectory optimization.
While these methods offer stronger safety assurances, they overlook user preferences, causing discomfort and reduced trust in ADS \cite{basu2017you}.
We leverage an optimization-based approach that formulates user preferences as constraints to optimize trajectories in the user preference space while ensuring collision-avoidance with obstacles.
In addition, we solve the optimization problems in parallel to generate possible trajectories in the driving scene.

\subsection{Preference-aware planning}
User preferences are influenced by various factors such as age\cite{chen2019driving}, personality traits\cite{hang2020human}, social disposition\cite{schwarting2019social, toghi2022social}, and psychological responses\cite{hu2023formulating}. 
For example, drivers with altruistic tendencies may consistently prefer to yield to surrounding vehicles, even if it means compromising their own driving goals. 
Furthermore, the presence of large vehicles, often perceived as threatening, can lead drivers to favor more conservative trajectories \cite{hu2023formulating, hu2024uncertainty}.
Incorporating such preferences into planning is crucial, as mismatches between the vehicle behavior and the user expectation may cause discomfort and reduce trust in ADS \cite{basu2017you}.
To address this limitation, recent studies have reflected preferences.

\textit{Preference-aware planning} \cite{wang2018driving, huang2021personalized, wang2022gaussian, wen2023modeling, wang2017driving, kim2021driving, schrum2024maveric} focuses on incorporating user preferences through unsupervised learning \cite{wang2018driving, huang2021personalized, wang2022gaussian, wen2023modeling} and supervised learning \cite{wang2017driving, kim2021driving, schrum2024maveric}. 
Unsupervised learning approaches typically extract preference-related features from user driving data, for example, by clustering primitive driving patterns \cite{wang2018driving} or inferring reward functions from user behavior \cite{wen2023modeling}. 
Supervised learning approaches use labeled or partially labeled driving style data to train models that classify styles and adjust trajectories. 
For example, \cite{schrum2024maveric} combines user driving data in simulation with questionnaire responses to identify users' driving style features and modulate the aggressiveness of the planned trajectory.
All of these approaches aim to identify implicit features to distinguish driving styles.
However, uncovering such features requires a diverse dataset that includes high-level factors such as drivers' psychological traits and social characteristics.
Labeling such factors to assemble a large dataset for learning is challenging as self-reported driving styles may differ from drivers' actual behavior \cite{basu2017you}.
We address this challenge by obtaining user-preferred clearance through a questionnaire.
We then construct the preference space using the questionnaire responses to explicitly incorporate user preferences into OCP by using this space as constraints.

%% file: Problem_statement.tex
\section{Method}

In this section, we describe the design of the proposed preference-aware planning framework. 
First, we design a questionnaire to obtain the user-preferred clearance margin, representing the minimum acceptable longitudinal distance that users prefer to maintain. 
Then, we formulate a trajectory optimization method using the preference space constructed from the questionnaire responses.

\subsection{Driving environment setup}
\begin{figure}[t]
\captionsetup{singlelinecheck = false, font=small, labelfont=bf, skip=0pt}
\captionsetup[subfloat]{singlelinecheck = true, labelfont=scriptsize,textfont=scriptsize}
\centering
\vspace{5pt}
\includegraphics[height=2.5cm]{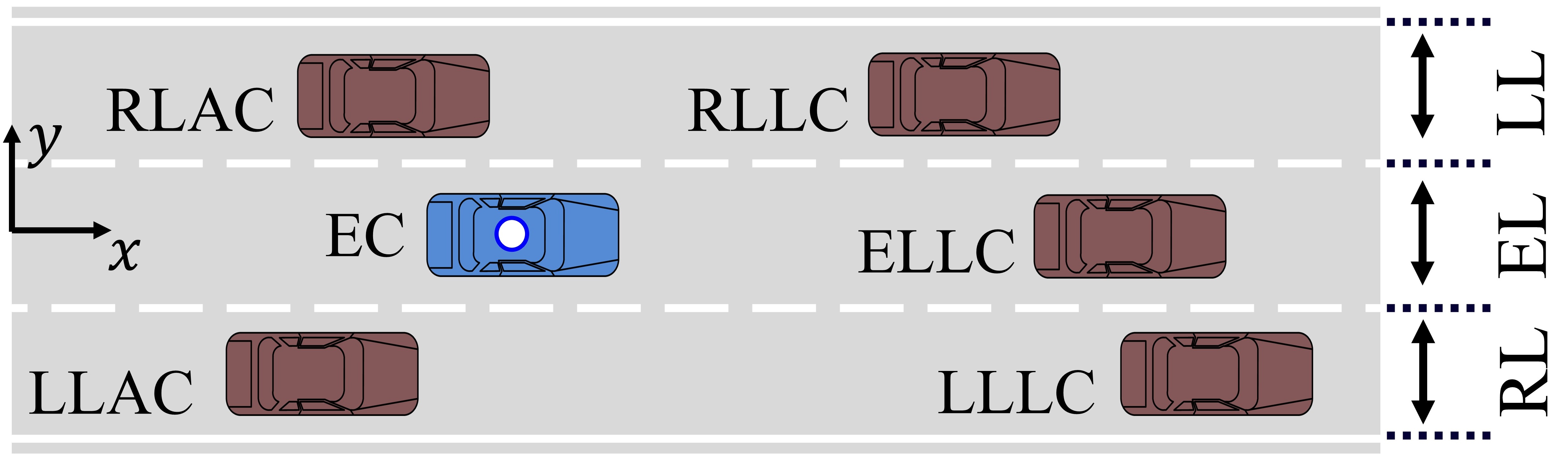}
\caption{
The $x$- and $y$-coordinates are equal to the longitudinal and lateral coordinates, respectively, because the Frenet frame represents curved roads as straight \cite{werling2010optimal}.
The abbreviations are summarized in Table~\ref{tab:abb}.
}
\label{fig:problem_statement}
\vspace{-5pt}
\end{figure}

We define vehicles that may influence the motion of the ego car (EC) as \textit{target vehicles}. 
As illustrated in Fig.~\ref{fig:problem_statement}, we consider five target vehicles \cite{kesting2007general, shalev2017formal} for obtaining the user-preferred clearance margin.
The driving environment has no traffic signals, and lanes are treated as straight using the Frenet frame that represents curved roads as straight \cite{werling2010optimal}.
In addition, we restrict EC to speeds below 60 kph, since high-speed driving involves complex vehicle dynamics.

Vehicles exhibit various states and maneuvers while driving on the road.
These are represented using the abbreviations provided in Table~\ref{tab:abb}.
We classify the maneuver type of the target vehicles based on the vehicle's heading angle \( \psi \):
\begin{equation}
\text{Maneuver} = \text{LLC if } \psi \geq 8^\circ,\ \text{RLC if } \psi \leq -8^\circ,\ \text{else LK.}
\label{equ:maneuver}
\end{equation}
From the maneuver type, TL of the vehicle is determined.
For example, if the vehicle is performing RLC, TL is set to the lane immediately to the right of the current lane. 
We also classify the size of the target vehicles:
\begin{equation}
\text{Size} = \textit{big} \text{ if } l \geq 7\,\text{m, else }\textit{small}\text{,}
\label{equ:size}
\end{equation}
where $l$ denotes the vehicle length.
In addition, the target vehicles are assigned to each lane (LL, RL, EL) based on their current lateral positions.
Finally, we predict the target vehicle's motion using a constant velocity model, allowing only single lane changes per transition.

\begin{table}[t]
\centering
\caption{Abbreviations}
\label{tab:abb}
\begin{tabular}{l l}
\hline
\textbf{Symbol} & \textbf{Description} \\
\hline
EC & Ego Car (controlled by the user) \\
LC & Leading Car (closest ahead of EC) \\
AC & Adjacent Car (closest behind EC) \\
LK, LLC, RLC & Lane Keeping, Left and Right Lane Change \\
EL, LL, RL & Ego lane, Left lane, Right lane \\
TL & Target Lane (by maneuver) \\
RLLC, ELLC, LLLC & Leading car in RL, EL, and LL \\
RLAC, LLAC & Adjacent car in RL and LL \\
TL of EC / LC / AC & Target lane by maneuver of EC / LC / AC \\
\hline
\end{tabular}
\vspace{-5pt}
\end{table}

\subsection{Design of questionnaire}
The current driving scenario, $\mathcal{D}=\mathcal{O}\times \boldsymbol{x}_{EC}$, is constructed from the target vehicle set $\mathcal{O} = $$\{\boldsymbol{x}_{\text{RLAC}}$, $\boldsymbol{x}_{\text{LLAC}}$, $\boldsymbol{x}_{\text{RLLC}}$, $\boldsymbol{x}_{\text{LLLC}}$, $\boldsymbol{x}_{\text{ELLC}}\}$ and EC state $\boldsymbol{x}_\text{EC}$.
We decide whether $\mathcal{D}$ is preferred using 
a preference function $\mathcal{P}: \mathcal{D} \rightarrow \mathbb{Z}_{\geq 0}$, where $\mathcal{P}(\mathcal{D}) = 0$ indicates a preferred scenario, else an unpreferred.
We assume that $\mathcal{P}(\cdot)$ is the combination of the interactions between a single-vehicle and EC:
\begin{equation}
\mathcal{P}(\mathcal{D}) = \sum_{i=1}^{n(\mathcal{O})} p\left(\mathcal{O}^{(i)}, \boldsymbol{x}_{\text{EC}}\right),
\label{equ:pref_decompose}
\end{equation}
where $\mathcal{O}^{(i)}$ is the $i$-th member of $\mathcal{O}$.
$p: \mathcal{O}^{(i)} \times \boldsymbol{x}_{\text{EC}} \rightarrow \{0, 1\}$ is a binary function that indicates whether the relation between $\boldsymbol{x}_{\text{EC}}$ and $\mathcal{O}^{(i)}$ is preferred, where $p(\mathcal{O}^{(i)}, \boldsymbol{x}_{\text{EC}}) = 1$ denotes the unpreferred case.
Our questionnaire uses \eqref{equ:pref_decompose}, thereby enabling the representation of the entire scenario space through single-vehicle interactions.

\begin{figure}[t]
\captionsetup{singlelinecheck = false, font=small, labelfont=bf, skip=0pt}
\captionsetup[subfloat]{singlelinecheck = true, labelfont=scriptsize,textfont=scriptsize}
\centering
\vspace{5pt}
\includegraphics[height=4.5cm]{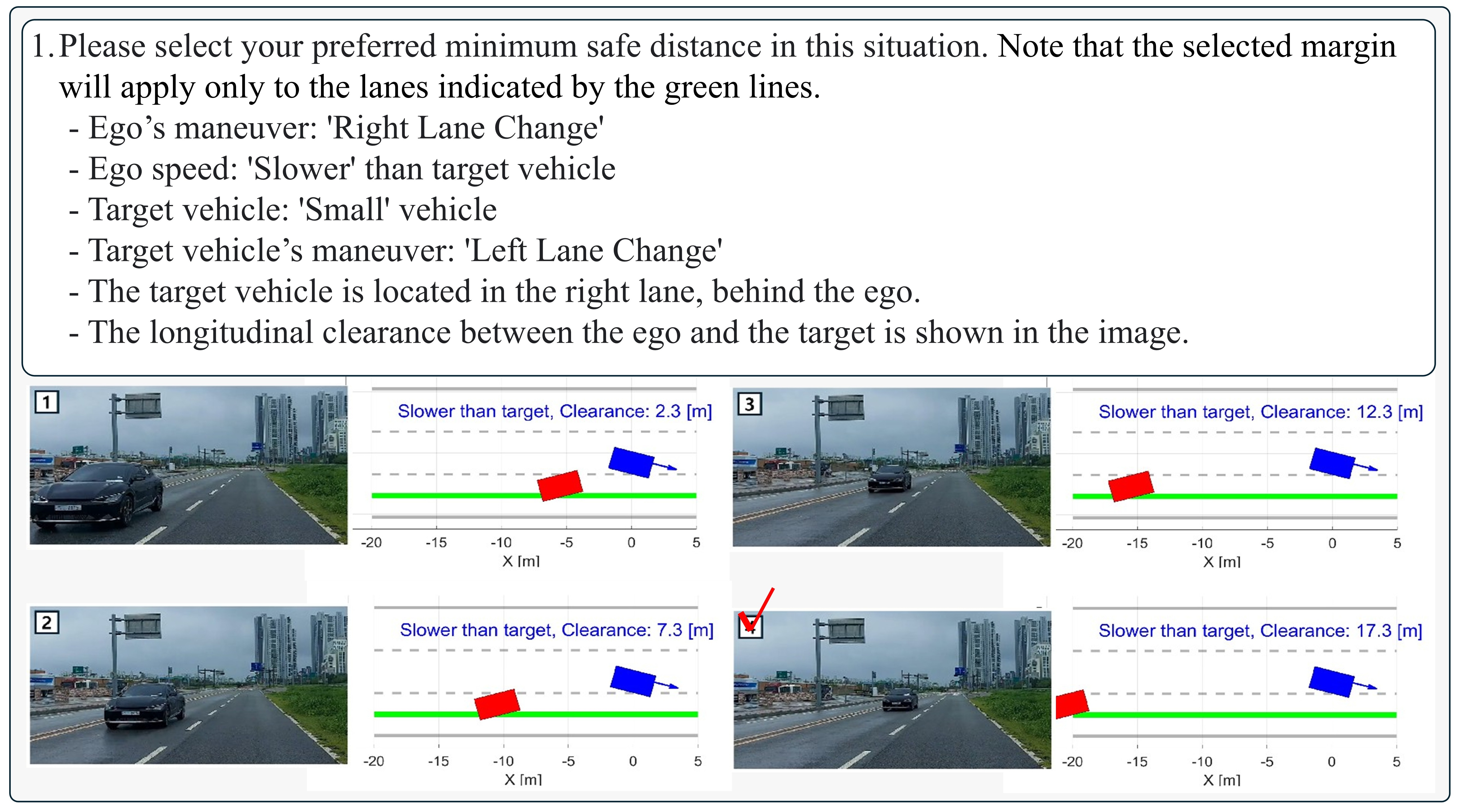}
\caption{
An example of a question in the questionnaire to address the single-vehicle interactions.
}
\label{fig:questionnaire}
\vspace{-10pt}
\end{figure}

We represent the single-vehicle interaction $p(\cdot)$ as the question $q \in Q$, where $Q$ denotes the questionnaire.
To define $Q$ as a finite set, we discretize $\mathcal{O}^{(i)}$ relative to $\boldsymbol{x}_{\text{EC}}$ based on human discrete perception \cite{freedman2008neural}.
\begin{equation}
Q=\{q\;|\;q\in\mathcal{T_M} \times \mathcal{E_M} \times \mathcal{B} \times \mathcal{F} \times \mathcal{L} \times \mathcal{R}\},
\label{equ:scenario_attribute}
\end{equation}
where $\mathcal{T_M}$ and $\mathcal{E_M}$ mean the maneuver of the target vehicle and EC, respectively.
$\mathcal{B}$, $\mathcal{F}$, $\mathcal{L}$, and $\mathcal{R}$ represent the discretized state of each target vehicle relative to EC.
\begin{enumerate}[noitemsep, topsep=0pt, leftmargin=*]
\item $\textit{Target Maneuver } \mathcal{T_M} = \{\text{LLC}, \text{RLC}, \text{LK} \}$ in \eqref{equ:maneuver}.

\item $\textit{Ego Maneuver } \mathcal{E_M}= \{\text{LLC}, \text{RLC}, \text{LK} \}$ in \eqref{equ:maneuver}.

\item \textit{Size set} $\mathcal{B} = \{\textit{small}, \textit{big}\}$ in \eqref{equ:size}.
    
\item \textit{Speed set} $\mathcal{F} = \{\textit{slower}, \textit{ faster}\}$, 
where the velocity of EC is compared with the target vehicle to determine whether EC is \textit{slower} or \textit{faster}.

\item \textit{Lane set} $\mathcal{L} = \{\text{LL}, \text{EL}, \text{RL} \}$. The target vehicle is assigned to a specific lane based on its lateral position.

\item \textit{Relative position set} $\mathcal{R}= \{\textit{rear}, \textit{front}\}$. 
The target vehicle is either behind (\textit{rear}) or in front of EC (\textit{front}).
\end{enumerate}
We denote these discrete representations as \textit{scenario attributes}.
Using the scenario attributes, we extract the user-preferred clearance margins for each target vehicle from the responses provided in the questionnaire.
As the choice in the questionnaire, we define the \textit{Discretized longitudinal position}, representing the relative longitudinal position of $\mathcal{O}^{(i)}$ from $\boldsymbol{x}_{\text{EC}}$.
The relative longitudinal position of $\mathcal{O}^{(i)}$ is partitioned into discrete intervals of 5 m.

We also include the response option \textit{“I don't care”} in the question $q$, allowing the user to indicate that the scenario defined in the question does not affect their driving preference. 
The full questionnaire is available \href{https://docs.google.com/forms/d/e/1FAIpQLSfFUCmlLn1dzrp17jE-ugcUBpI16Ywin6hH7Mu2zTBXsjOh2A/viewform?usp=dialog}{here} and an example of the question is shown in Fig.~\ref{fig:questionnaire}.
Through the questionnaire responses, we identify the user-preferred clearance margin for the single-vehicle interaction between $\mathcal{O}^{(i)}$ and $\boldsymbol{x}_{\text{EC}}$ described in each question $q$.

%% file: method.tex
\begin{table}[!b]
\centering
\caption{List of low-level strategies ($\Phi$)}
\label{tab:low_level_strategies}
\setlength{\tabcolsep}{3pt}
\begin{tabular}{c  c  c  c}
\hline
\textbf{ID} & ($\phi$) \textbf{Low-level strategy} & ${v_\mathrm{des}}$ & ${y_\mathrm{des}}$ \\ 
\hline
1  & RLC ahead of RLLC & $v_\text{f}$ &$ f_\text{c}(\text{RL})$\\
\hline
2  & RLC ahead of RLAC and RLLC & $v_\text{f}$ &$ f_\text{c}(\text{RL})$\\
\hline
3  & RLC ahead of RLAC and behind RLLC & $v_\text{f}$ & $f_\text{c}(\text{RL})$\\
\hline
4  & RLC behind RLLC & $v_\text{f}$ & $f_\text{c}(\text{RL})$\\
\hline
5  & RLC behind RLLC and RLAC & $v_\text{f}$ & $f_\text{c}(\text{RL})$\\
\hline
6  & RLC ahead of RLAC & $v_\text{f}$ & $f_\text{c}(\text{RL})$\\
\hline
7  & RLC behind RLAC & $v_\text{f}$ & $f_\text{c}(\text{RL})$\\
\hline
8  & RLC without target vehicles at high speed & $v_\text{f}$ & $f_\text{c}(\text{RL})$\\
\hline
9  & RLC without target vehicles at moderate speed & $v_\text{m}$ & $f_\text{c}(\text{RL})$\\
\hline
10  & RLC without target vehicles at slow speed & $v_\text{s}$ & $f_\text{c}(\text{RL})$\\
\hline
11  & LLC ahead of RLLC & $v_\text{f}$ &$ f_\text{c}(\text{LL})$\\
\hline
12  & LLC ahead of RLAC and RLLC & $v_\text{f}$ &$ f_\text{c}(\text{LL})$\\
\hline
13  & LLC ahead of RLAC and behind RLLC & $v_\text{f}$ & $f_\text{c}(\text{LL})$\\
\hline
14  & LLC behind RLLC & $v_\text{f}$ & $f_\text{c}(\text{LL})$\\
\hline
15  & LLC behind RLLC and RLAC & $v_\text{f}$ & $f_\text{c}(\text{LL})$\\
\hline
16  & LLC ahead of RLAC & $v_\text{f}$ & $f_\text{c}(\text{LL})$\\
\hline
17  & LLC behind RLAC & $v_\text{f}$ & $f_\text{c}(\text{LL})$\\
\hline
18  & LLC without target vehicles at high speed & $v_\text{f}$ & $f_\text{c}(\text{LL})$\\
\hline
19  & LLC without target vehicles at moderate speed & $v_\text{m}$ & $f_\text{c}(\text{LL})$\\
\hline
20  & LLC without target vehicles at slow speed & $v_\text{s}$ & $f_\text{c}(\text{LL})$\\
\hline
21 & LK faster than RLLC and LLLC & $v_\text{f}$ & $f_\text{c}(\text{EL})$\\
\hline
22 & LK faster than RLLC and slower than LLLC & $v_\text{m}$ & $f_\text{c}(\text{EL})$\\
\hline
23 & LK faster than LLLC and slower than RLLC & $v_\text{m}$ &  $f_\text{c}(\text{EL})$\\
\hline
24 & LK slower than RLLC and LLLC & $v_\text{s}$ & $f_\text{c}(\text{EL})$\\
\hline
25 & LK faster than RLLC & $v_\text{f}$ & $f_\text{c}(\text{EL})$\\
\hline
26 & LK slower than RLLC & $v_\text{m}$ & $f_\text{c}(\text{EL})$\\
\hline
27 & LK faster than LLLC & $v_\text{f}$& $f_\text{c}(\text{EL})$\\
\hline
28 & LK slower than LLLC & $v_\text{m}$& $f_\text{c}(\text{EL})$\\
\hline
29 & LK without RLLC and LLLC at high speed & $v_\text{f}$ & $f_\text{c}(\text{EL})$\\
\hline
30 & LK without RLLC and LLLC at moderate speed & $v_\text{m}$ & $f_\text{c}(\text{EL})$\\
\hline
31 & LK without RLLC and LLLC at slow speed & $v_\text{s}$ & $f_\text{c}(\text{EL})$\\
\hline
32 & LK stop & 0 & $f_\text{c}(\text{EL})$\\ \hline
\end{tabular}

\begin{flushleft}
\justifying
\footnotesize
$v_\text{f}$, $v_\text{m}$, and $v_\text{s}$ represent the fast, moderate, and slow velocities, respectively.
$f_\text{c}(\text{EL})$, $f_\text{c}(\text{RL})$, and $f_\text{c}(\text{LL})$ indicate the lane center of EL, RL, and LL, respectively.
Note that each low-level strategy is only applicable under the driving scene where only the target vehicles explicitly mentioned in the strategy are present.
For example, the strategy "RLC ahead of RLLC" (ID 1) is valid when RLLC is the sole vehicle present in RL.
\end{flushleft}
\vspace{-10pt}
\end{table}

\subsection{Optimization-based preference-aware planning}
As shown in Fig.~\ref{fig:framework}, our planner starts with the user answering the questionnaire.
The collected responses are used to construct a user preference space, representing the user-preferred region regarding the target vehicles.
This space is then incorporated as preference constraints in the OCP for preference-aware planning:
\begin{equation}
\begin{aligned}
\min_{\{\boldsymbol{u}[k]\}_{k=0}^{N-1}} \quad &J(\boldsymbol{x}[k], \boldsymbol{u}[k]) \\
\text{s.t.} \quad & \boldsymbol{x}[k+1] = f_\text{sys}(\boldsymbol{x}[k], \boldsymbol{u}[k]), \\
& \boldsymbol{x}[k] \in \mathcal{X}, \boldsymbol{u}[k] \in \mathcal{U}, \\
& h_i(\mathcal{O}, \boldsymbol{x}[k]) \times f_i(\boldsymbol{x}[k]) \geq 0, \forall i=1,...,M,
\end{aligned}
\label{equ:preference_opt}
\end{equation}
where $J$ represents the cost function.
$\boldsymbol{x}[k] \in \mathbb{R}^n$ and $\boldsymbol{u}[k] \in \mathbb{R}^m$ denote the state and control input at step $k$, respectively. 
The system dynamics is given by $f_\text{sys}(\cdot)$, and $\mathcal{X}$ and $\mathcal{U}$ denote the feasible state and control input spaces, respectively. 
The details about state, control input, system dynamics, $\mathcal{X}$, and $\mathcal{U}$ are provided in Sec.~\ref{sec:sim_result}.
$M$ is the total number of scenarios that can be generated in $\mathcal{D}$ over the planning horizon $N$.
$f_i$ is obtained by the user-preferred margins obtained from $Q$.
$h_i$ is the condition function that determines whether the corresponding scenario is active when $h_i = 1$, else $h_i = 0$.
Since $q \in Q$ is based on a specific scenario and $h_i$ describes the corresponding scenario, the preference constraints $f_i$ are conditioned on scenario attributes in \eqref{equ:scenario_attribute}.

Among these attributes, $\mathcal{F}$, $\mathcal{E_M}$, $\mathcal{L}$, and $\mathcal{R}$ are related to movement of EC; thus they can change over the planning horizon in \eqref{equ:preference_opt}.
Whereas \( \mathcal{B} \) and \( \mathcal{T_M} \) are determined in advance from \eqref{equ:size} and \eqref{equ:maneuver}, they remain fixed over the planning horizon.
Since the number of distinct scenarios in \eqref{equ:preference_opt} depends on the combinations of $\mathcal{F}$, $\mathcal{E_M}$, $\mathcal{L}$, and $\mathcal{R}$, it calculates as $M=(n(\mathcal{F})\times n(\mathcal{E_M})\times n(\mathcal{L})\times n(\mathcal{R}))^N$, making real-time implementation intractable.

To address this limitation, we decompose \eqref{equ:preference_opt} into scenario-specific subproblems.
For each subproblem, we fix the scenario attributes of each target vehicle to ensure that \(f_i\) becomes an unconditioned constraint.
We denote this \textit{low-level strategy} $\phi$.
The subproblem with $\phi$ is
\begin{equation}
\begin{aligned}
\min_{\{\boldsymbol{u}[k]\}_{k=0}^{N-1}} \quad & \sum_{k=0}^{N} V\left(\boldsymbol{x}[k], \boldsymbol{u}[k]\right) \\
\text{s.t.} \quad & \boldsymbol{x}[k+1] = f_\text{sys}(\boldsymbol{x}[k], \boldsymbol{u}[k]), \\
& \boldsymbol{x}[k] \in \mathcal{S_\phi} \cap \mathcal{X},\; \boldsymbol{u}[k] \in \mathcal{U}, \\
\end{aligned}
\label{equ:app_preference_opt}
\end{equation}
where $V$ denotes the scenario-specific cost function, and $\mathcal{S_\phi}$ denotes the \textit{preference-based free space} constructed based on the response from $Q$.
We set $V$ in \eqref{equ:app_preference_opt} as
\begin{equation}
    V\left(\boldsymbol{x}[k]\right) = W_1(y[k]-y_\text{des}[k])^2 +  W_2(v[k]-v_\text{des}[k])^2, \nonumber
\end{equation}
\noindent where $y_\text{des}$ and $v_\text{des}$ are the desired lateral position and the desired velocity, respectively, from $\phi$.
\textcolor{black}{
$y[k]$ and $v[k]$ represent the lateral position and velocity at step $k$, respectively.}
$W_1$ and $W_2$ denote the corresponding weights.
The list of available $\phi$, $y_\text{des}$, and $v_\text{des}$ is summarized in Table~\ref{tab:low_level_strategies}.

\begin{figure}[t]
\captionsetup{singlelinecheck = false, font=small, labelfont=bf, skip=-5pt}
\captionsetup[subfloat]{singlelinecheck = true, labelfont=scriptsize,textfont=scriptsize}
\centering
\vspace{5pt}
\includegraphics[height=4.0cm]{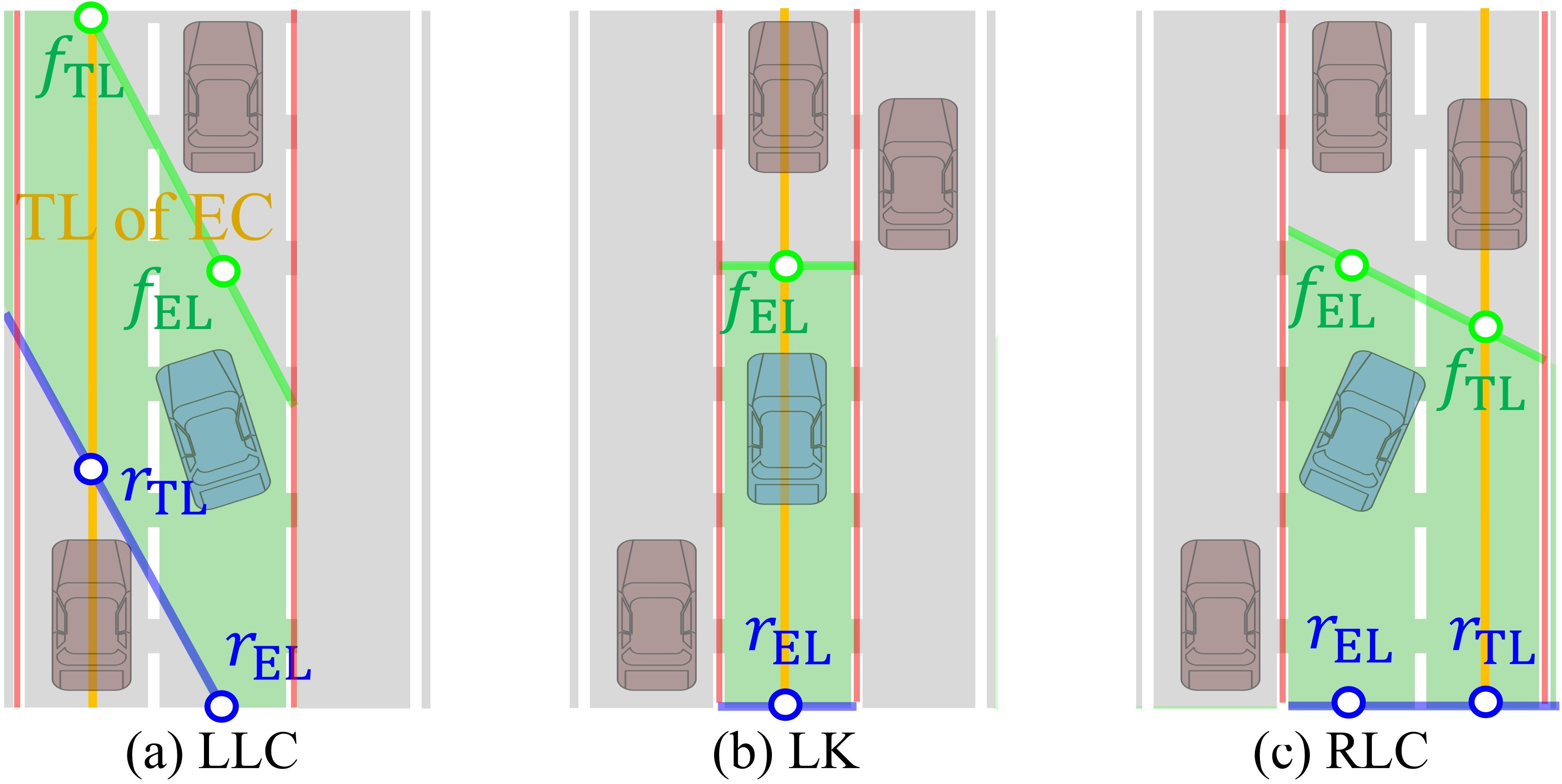}
\caption{The blue, green, and red solid lines indicate the lower and upper bounds of the preference space.
The green shaded region means the preference space.
(a), (b), and (c) show the preference spaces for the LLC, LK, and RLC cases, respectively. 
Each preference space is constructed based on user responses to the questionnaire.  
}
\label{fig:preference_space}
\vspace{-5pt}
\end{figure}

\begin{figure}[t]
\captionsetup{singlelinecheck = false, font=small, labelfont=bf, skip=0pt}
\captionsetup[subfloat]{singlelinecheck = true, labelfont=scriptsize,textfont=scriptsize}
\centering
\vspace{5pt}
\includegraphics[height=2.5cm]{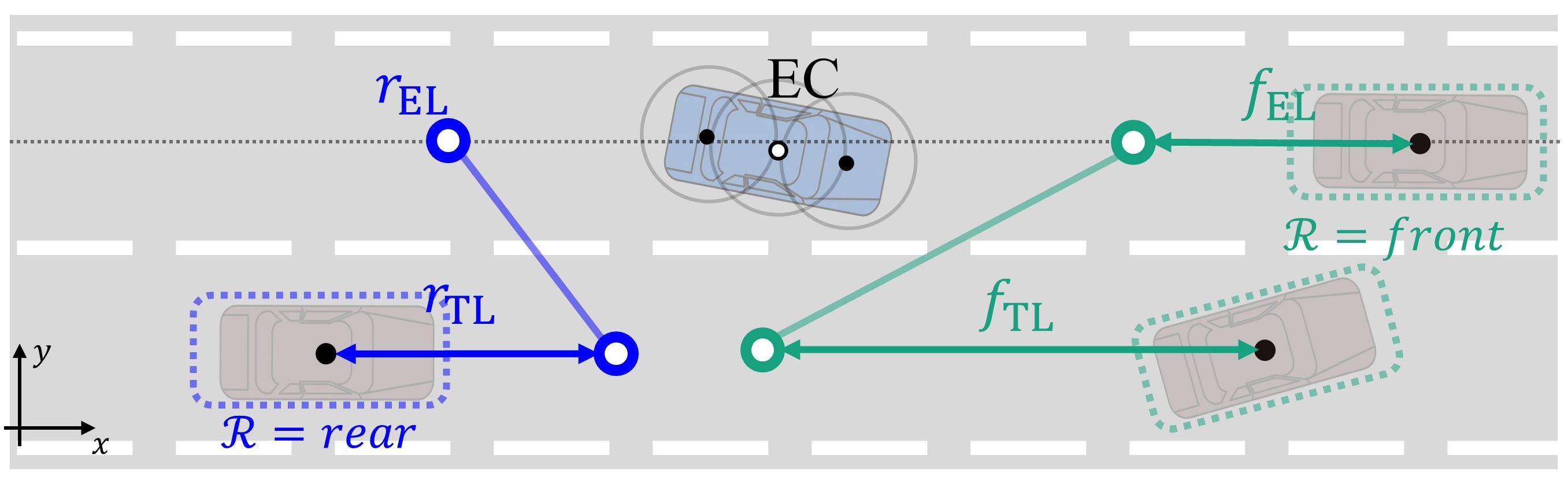}
\caption{
We refine $r_\text{EL}$, $r_\text{TL}$, $f_\text{EL}$, and $f_\text{TL}$ using a separating hyperplane, considering the size of both the target vehicles and EC.
}
\label{fig:safe_point}
\vspace{-5pt}
\end{figure}

As shown in Fig. \ref{fig:preference_space}, $r_\text{EL}$, $r_\text{TL}$, $f_\text{EL}$, and $f_\text{TL}$ determine the preference space.
$r_\text{EL}$, $r_\text{TL}$, $f_\text{EL}$, and $f_\text{TL}$ are each represented as an ($x, y$) pair, where $x$ denotes the user-preferred clearance margin and $y$ corresponds to the $y$-position of the vehicle that defines the respective margin.
The process of obtaining $r_\text{EL}$, $r_\text{TL}$, $f_\text{EL}$, and $f_\text{TL}$ is described in Alg.~\ref{alg:pref}.
We first select the available $\phi$ in $\mathcal{D}$ from Table~\ref{tab:low_level_strategies}. 
From $\phi$, we determine $\mathcal{E_M}$, TL, and EL.
The lateral positions of $r_\text{EL}$ and $f_\text{EL}$ are aligned with the center of EL, while those of $r_\text{TL}$ and $f_\text{TL}$ are aligned with the center of TL of EC.
To handle cases with no target vehicles, we assign default values $r_\text{EL}$, $r_\text{TL}$, $f_\text{EL}$, and $f_\text{TL}$ (line 1).
By default, $r_\text{EL}$ and $r_\text{TL}$ are set to $-30\,\mathrm{m}$ and $f_\text{EL}$ and $f_\text{TL}$ to $80\,\mathrm{m}$, relative to EC's $x$ position along the longitudinal direction.
Finally, we assign scenario attributes to each vehicle based on the present state of EC. 
Then, using the responses to the questionnaire corresponding to these attributes, we determine $r_\text{EL}$, $r_\text{TL}$, $f_\text{EL}$, and $f_\text{TL}$ (line 2-14). 
However, the calculated preference space using $r_\text{EL}$, $r_\text{TL}$, $f_\text{EL}$, and $f_\text{TL}$ is not always within the free space, as the line ((\( r_\text{EL} \) and \(  r_\text{TL} \)) or (\( f_\text{EL} \) and \( f_\text{TL} \))) may cross into the regions occupied by the target vehicles, as shown in Fig.~\ref{fig:preference_space}(b-c). 
To address this, we refine $r_\text{EL}$, $r_\text{TL}$, $f_\text{EL}$, and $f_\text{TL}$ using the separating hyperplanes between the target vehicles and EC, as shown in Fig.~\ref{fig:safe_point}.
In each region divided by the hyperplane, the area occupied by the obstacle is defined to include the original $r_\text{EL}$, $r_\text{TL}$, $f_\text{EL}$, and $f_\text{TL}$.
Finally, we obtain the preference-based free space $\mathcal{S}_\phi$ in \eqref{equ:app_preference_opt}.

\begin{algorithm}[!b]
\caption{Constructing preference space}
\label{alg:pref}
\begin{algorithmic}[1]
\STATE \textbf{Input}: $\phi$, $\mathcal{D}$, $\mathcal{E_M}$, TL, EL
\STATE \textbf{Output}: $r_\text{EL},r_\text{TL},f_\text{EL},f_\text{TL}$
\STATE
\STATE $r_\text{EL},r_\text{TL},f_\text{EL},f_\text{TL} \gets \textbf{setDefault}(\mathcal{D}, \mathcal{E_M})$
\STATE \textbf{for} $X_\text{veh}$ \textbf{in} $\mathcal{O}$:
\STATE \hspace{0.3cm} \textbf{if} $X_{\text{veh}}$ exists in EL \textbf{then}
\STATE \hspace{0.6cm} $\left[\mathcal{F}, \mathcal{B}, \mathcal{T_M},\mathcal{L},\mathcal{R}\right] \gets \textbf{vehicleInfo}\left(X_\text{veh}\right)$
\STATE \hspace{0.6cm} \textbf{if} $X_{\text{veh}}$ is LC \textbf{then}
\STATE \hspace{0.9cm} $f_\text{EL} \gets \textbf{questionnaire}\left(\mathcal{B}, \mathcal{F}, \mathcal{T_M}, \mathcal{E_M}, \mathcal{L}, \mathcal{R}\right)$
\STATE \hspace{0.6cm} $\textbf{else}$
\STATE \hspace{0.9cm} $r_\text{EL} \gets \textbf{questionnaire}\left(\mathcal{B}, \mathcal{F}, \mathcal{T_M}, \mathcal{E_M}, \mathcal{L}, \mathcal{R}\right)$

\STATE \hspace{0.3cm} \textbf{if} $X_{\text{veh}}$ exists in TL \textbf{then}
\STATE \hspace{0.6cm} $\left[\mathcal{F}, \mathcal{B}, \mathcal{T_M},\mathcal{L},\mathcal{R}\right] \gets \textbf{vehicleInfo}\left(X_\text{veh}\right)$
\STATE \hspace{0.6cm} \textbf{if} $X_{\text{veh}}$ is LC \textbf{then}
\STATE \hspace{0.9cm} $f_\text{TL} \gets \textbf{questionnaire}\left(\mathcal{B}, \mathcal{F}, \mathcal{T_M}, \mathcal{E_M}, \mathcal{L}, \mathcal{R}\right)$
\STATE \hspace{0.6cm} $\textbf{else}$
\STATE \hspace{0.9cm} $r_\text{TL} \gets \textbf{questionnaire}\left(\mathcal{B}, \mathcal{F}, \mathcal{T_M}, \mathcal{E_M}, \mathcal{L}, \mathcal{R}\right)$
\end{algorithmic}
\end{algorithm}

After constructing $S_\phi$ and solving \eqref{equ:app_preference_opt} for all available $\phi \in \Phi$, the approximate solution of \eqref{equ:preference_opt} is obtained by selecting the trajectory that minimizes the original cost function.
\begin{equation}
\tau_p^*=\argmin_{\tau_\phi\in\Gamma}J\left(\tau_\phi\right), \nonumber
\label{equ:selection}
\end{equation}
where $\tau_p^*$ means the user-preferred trajectory, and $\tau_\phi$ is the solution of the subproblem~\eqref{equ:app_preference_opt} corresponding to $\phi$.
$\Gamma$ denotes the solution set comprising the solutions of all \eqref{equ:app_preference_opt}.
In this study, we define $J$ for minimum time planning as follows:
\begin{equation}
J=     \sum_{k=0}^N (v[k]-v_\text{drive})^2,
\label{equ:cost}
\end{equation}
where we set $v_\text{drive} = 60$ kph as the maximum velocity used in this study.

%% file: Result.tex
\setlength{\textfloatsep}{15pt} 
\begin{table}[!b]
\centering
\caption{Parameter Setting.} 
\label{tab:hyperparameters}
\centering
\begin{tabular}{c|c|c|c}
\hline
 \textbf{Parameter} & \textbf{Description} & \textbf{Value} & \textbf{Unit}\\
\hline
$N$ & Horizon length & 60 & $\cdot$\\
\hline

$\Delta t$ & Time interval & 0.05 & [s] \\
\hline

$L$ & Vehicle wheelbase & 2.7 & [m]\\
\hline

$v_\text{drive}$ & Driving velocity & 60 & [kph]\\
\hline
$v_\text{f}$ & Desired velocity for $\phi$ & 60 & [kph]\\
\hline
$v_\text{m}$ & Desired velocity for $\phi$  & 45 & [kph]\\
\hline
$v_\text{s}$ & Desired velocity for $\phi$  & 30 & [kph]\\
\hline

$V_\text{max}$ & Maximum velocity & 60 & [kph]\\
\hline
$V_\text{min}$ & Minimum velocity & 0 & [kph]\\
\hline

$\psi_\text{max}$ & Maximum heading angle & 12 & [deg]\\
\hline
$\psi_\text{min}$ & Minimum heading angle & -12 & [deg]\\
\hline

$a_\text{max}$ & Maximum acceleration & 2 & [m/s$^2$]\\
\hline
$a_\text{min}$ & Minimum acceleration & -4 & [m/s$^2$]\\
\hline

$j_\text{max}$ & Maximum jerk & 6 & [m/s$^3$]\\
\hline
$j_\text{min}$ & Minimum jerk & -6 & [m/s$^3$]\\
\hline

$\delta_\text{max}$ & Maximum steering angle & 32 & [deg]\\
\hline
$\delta_\text{min}$ & Minimum steering angle & -32 & [deg]\\
\hline

$\gamma_\text{max}$ & Maximum steering angle rate & 0.346 & [rad/s]\\
\hline

$\gamma_\text{min}$ & Minimum steering angle rate & -0.346 & [rad/s]\\
\hline

$W_{1}$ & Weight of lateral state & 100 &\\
\hline
$W_{2}$ & Weight of velocity state & 1 &\\
\hline
\end{tabular}

\end{table}

\section{Simulation results}
\label{sec:sim_result}
For simulation, the system dynamics, state, and control input of EC are defined as follows:
\begin{equation}
\begin{aligned}
\boldsymbol{x} &= [x,\, y,\, v,\, \psi,\, a,\, \delta]^\top,\quad \boldsymbol{u} = [j,\, \gamma]^\top, \\
\dot{x} &= v \cos(\psi),\quad 
\dot{y} = v \sin(\psi),\quad 
\dot{v} = a, \\
\dot{\psi} &= \frac{v}{L} \tan(\delta),\quad
\dot{a} = j,\quad 
\dot{\delta} = \gamma,
\end{aligned}
\label{equ:system_dynamics}
\end{equation}
\noindent where $\boldsymbol{x}$ and $y$ represent the global position of the vehicle in Cartesian coordinates, while $v$ denotes its velocity. 
The heading angle is given by $\psi$, and $a$ corresponds to the longitudinal acceleration. 
$\delta$ indicates the steering angle, and $L$ stands for the length of the vehicle wheelbase. 
$j$ refers to the jerk and $\gamma$ describes the steering angle rate. 
Each component in $\boldsymbol{x}$ and $\boldsymbol{u}$ follows the activation constraints:
\begin{equation}
\begin{aligned}
V_\text{min} &\leq v \leq V_\text{max},
\;\; \psi_\text{min} \leq \psi \leq \psi_\text{max},
\;\; a_\text{min} \leq a \leq a_\text{max}, \nonumber \\
\delta_\text{min} &\leq \delta \leq \delta_\text{max},
\;\;\,\,\, j_\text{min} \leq j \leq j_\text{max},
\;\;\,\,\, \gamma_\text{min} \leq \gamma \leq \gamma_\text{max}, \nonumber \\
\end{aligned}
\end{equation}
where $V_\text{min}$, $\psi_\text{min}$, $a_\text{min}$, $\delta_\text{min}$, $j_\text{min}$, and $\gamma_\text{min}$ are the lower bounds of each corresponding variable, and 
$V_\text{max}$, $\psi_\text{max}$, $a_\text{max}$, $\delta_\text{max}$, $j_\text{max}$, and $\gamma_\text{max}$ are the upper bounds of each corresponding variable.
We linearize (\ref{equ:system_dynamics}) around the initial state $\boldsymbol{x}_0 = [x_0, y_0, v_0, \psi_0, a_0, \delta_0]^\top$ and discretize the linearized model using $\Delta t$ to use it as the dynamics constraint in \eqref{equ:preference_opt}.

\begin{figure}[!t]
\captionsetup{singlelinecheck = false, font=small, labelfont=bf, skip=-2pt}
\captionsetup[subfloat]{singlelinecheck = true, labelfont=scriptsize,textfont=scriptsize}
\centering
\vspace{5pt}
\includegraphics[height=7.0cm]{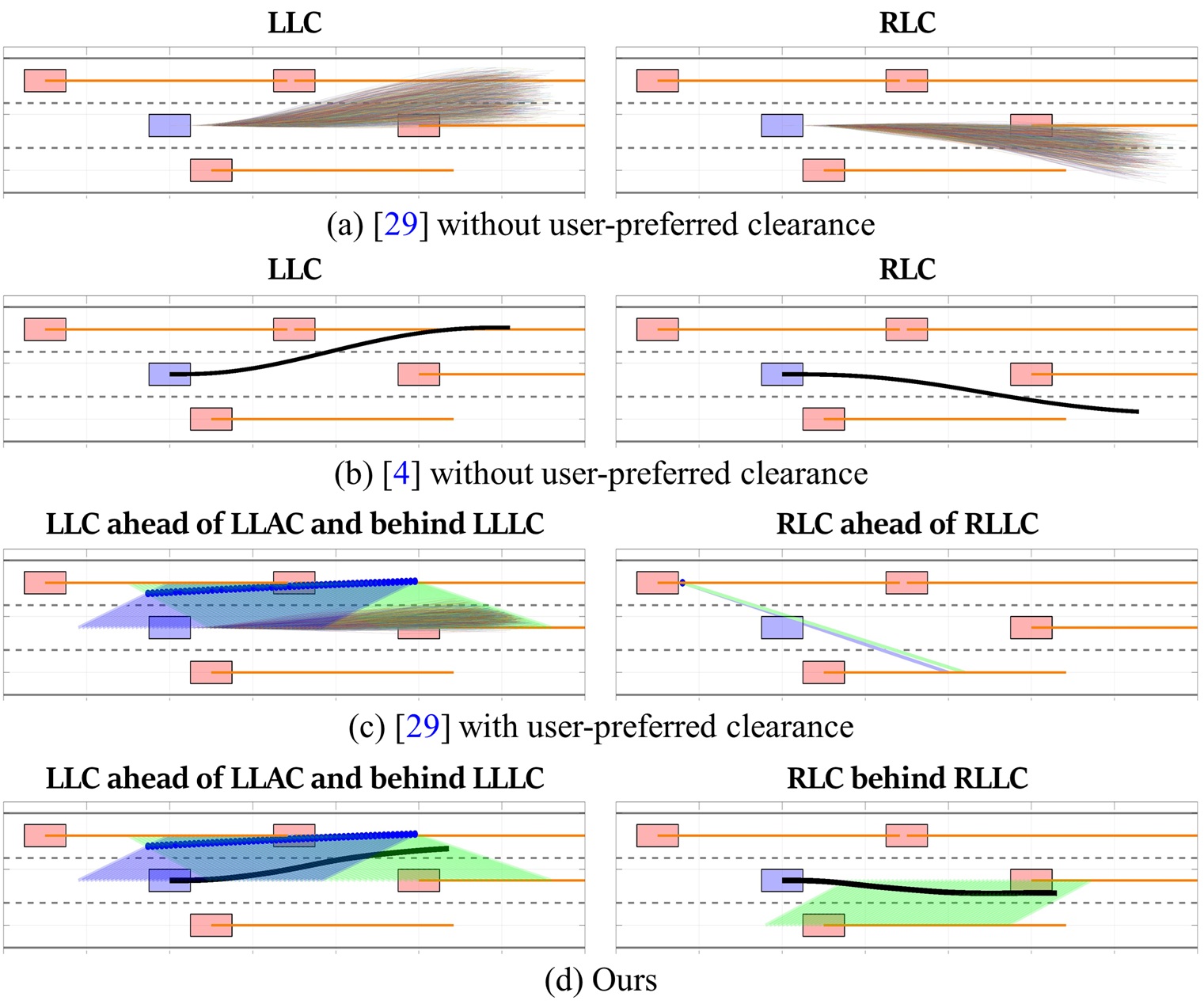}
\caption{ 
\textcolor{black}{
The {red} and {blue} boxes are the target vehicle and EC, respectively.
The {orange} and {black} solid lines mean the predicted trajectory of target vehicles and the optimized trajectory of EC, respectively, and the light blue and green lines in (c) and (d) mean the lower and upper bounds of $\mathcal{S}_\phi$.
The blue point in (c) and (d) represents the intersection point between the lower and upper bounds of $\mathcal{S}_\phi$.
The lines with various colors in (a) and (c) denote the trajectory samples.}
The trajectory samples in (c) and the optimized trajectories in (d) are restricted in $\mathcal{S}_\phi$ to reflect the user-preferred clearances.
When $\phi$ = RLC ahead of RLLC in (c), the current position of EC does not belong to $\mathcal{S}_\phi$; thus, there is no solution.
}
\label{fig:long-term_comparison}
\vspace{-0pt}
\end{figure}

The hyperparameters used for simulations are summarized in Table~\ref{tab:hyperparameters}.
We set the steering angle and acceleration limits considering the physical capabilities of the vehicle, while the steering angle change limit and jerk limit are set to ensure ride comfort \cite{huang2004fundamental}.
In addition, we limit the heading angle range to maintain the validity of the small-angle assumption, since we use the linearization of \eqref{equ:system_dynamics} for planning.
The time interval is set to 0.05 seconds, as our local planner operates at 20~Hz to enable rapid trajectory replanning.
The horizon length \(N\) is set to 60, corresponding to predictions up to 3 seconds into the future.
The maximum velocity \(V_{\max}\) is set to 60~kph to ensure the validity of the vehicle kinematics in \eqref{equ:system_dynamics}, as slipping is negligible below this speed.
We set $v_\mathrm{f}$, $v_\mathrm{m}$, and $v_\mathrm{s}$ to 60, 45, and 30~kph, respectively, for executing $\phi$ as listed in Table~\ref{tab:low_level_strategies}.

Before simulations, a person answers the questionnaire to generate a sample set of responses. 
Since it is possible to construct the preference-based free space regardless of the specific answers to the questionnaire, we assume a particular driver type (e.g., aggressive or conservative driver) and obtain corresponding responses. 
Based on the questionnaire, we compare our planner with the baselines by measuring the proportion of total driving time spent within the preference space.
This evaluation indicates how well the user-preferred clearances are reflected.

After generating the responses from the questionnaire, we evaluate our framework against preference-agnostic planners \cite{werling2010optimal, jin2023safety, adajania2022multi} to validate how well ours reflects the user-preferred clearances. 
Furthermore, we validate that our framework enables the vehicle to incorporate the user-preferred clearances across different driving styles.
Note that the preference-aware planning methods are excluded from the comparison because no publicly available datasets exist for fair evaluation \cite{wang2017driving, wen2023modeling, schrum2024maveric}.

\begin{figure}[!t]
\captionsetup{singlelinecheck = false, font=small, labelfont=bf, skip=-2pt}
\captionsetup[subfloat]{singlelinecheck = true, labelfont=scriptsize,textfont=scriptsize}
\centering
\vspace{5pt}
\includegraphics[height=1.9cm]{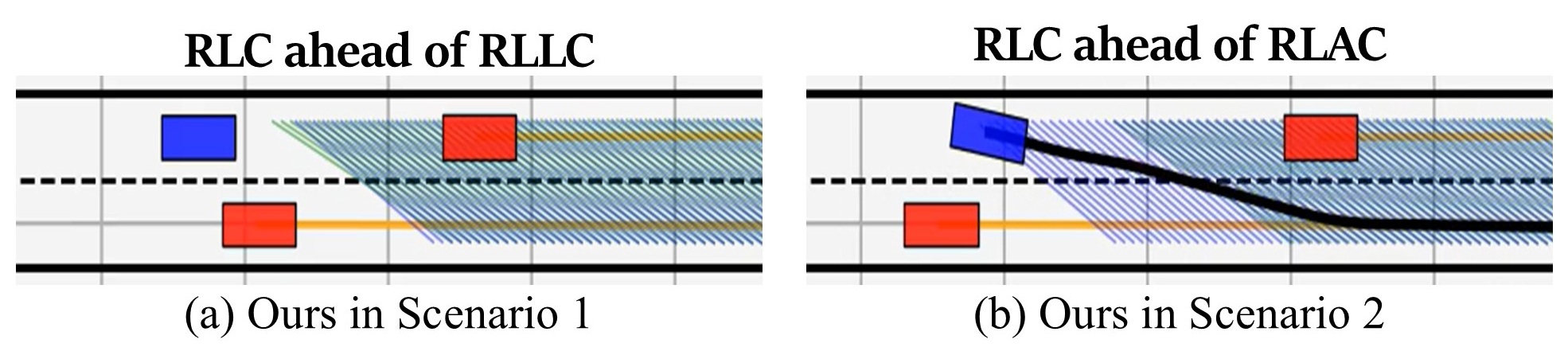}
\caption{
The {red} and {blue} boxes, the {orange} and {black} solid lines, and the light blue and green lines are the same as in Fig. \ref{fig:long-term_comparison}.
(a) and (b) depict the solution of the subproblem for overtaking.
The RLC ahead of RLLC in (a) does not have a feasible solution because the vehicle is not within $\mathcal{S}_\phi$. 
In contrast, the RLC ahead of RLAC in (b) has a feasible solution.
}
\label{fig:scenario_result}
\vspace{-5pt}
\end{figure}

\begin{table}[!b]
    \centering
    \caption{Results of driving 30 closed-loop simulations.}
    \small
    \setlength{\extrarowheight}{0pt}
    \renewcommand{\arraystretch}{1.2}  
    \begin{tabular}{l | c c c}
        \toprule
        \multicolumn{4}{>{\columncolor{gray!10}}c}{\textbf{Scenario 1}} \\
        \midrule
        \multirow{2}{*}{\textbf{Method}}  
        & {\textbf{Pref.}} & {\textbf{Avg.}} & {\textbf{Change}} \\
        & {\textbf{rate [\%]}} & {\textbf{speed [m/s]}} & {\textbf{count}} \\
        \midrule
        Frenet-Planner \cite{werling2010optimal} & --  & --  & -- \\
        Batch MPC \cite{adajania2022multi}      & 51.53  & 16.31  & 5.07 \\
        CBF + MPPI \cite{jin2023safety}         & 57.41  & 16.78  & 5.76 \\
        \textbf{Ours}                           & \textbf{100} & \textbf{12.50} & \textbf{0.47} \\
        \midrule
        \multicolumn{4}{>{\columncolor{gray!10}}c}{\textbf{Scenario 2}} \\
        \midrule
        \multirow{2}{*}{\textbf{Method}}  
        & {\textbf{Pref.}} & {\textbf{Avg.}} & {\textbf{Change}} \\
        & {\textbf{rate [\%]}} & {\textbf{speed [m/s]}} & {\textbf{count}} \\
        \midrule
        Frenet-Planner \cite{werling2010optimal} & 100.0  & 16.24 & 4.43 \\
        Batch MPC \cite{adajania2022multi}      & 89.59  & 16.45 & 4.50 \\
        CBF + MPPI \cite{jin2023safety}         & 89.11  & 16.52 & 4.53 \\
        \textbf{Ours}                           & \textbf{99.86} & \textbf{15.95} & \textbf{2.93} \\
        \bottomrule
    \end{tabular}
    \label{tab:long-term_comparison}
    
    \vspace{2pt}
    \begin{flushleft}
    \justifying
    \footnotesize
    Dash (-) indicates that the scenario is not completed and, therefore, the metric is not computed.  
    \textbf{Pref. rate} denotes the percentage of time the vehicle stayed within the preference space,  
    \textbf{Avg. speed} denotes the average driving speed, and  
    \textbf{Change count} denotes the average number of lane changes.
    \end{flushleft}
    \vspace{-10pt}
\end{table}

\subsection{Evaluation on preference-awareness}
We compare preference-agnostic planners \cite{jin2023safety, adajania2022multi, werling2010optimal} with our proposed framework.  
Furthermore, since \cite{jin2023safety} can be extended to incorporate user-preferred clearances, we also conduct a comparative study between our planning algorithm and \cite{jin2023safety} with our user-preferred clearances.
Fig.~\ref{fig:long-term_comparison} reports the qualitative comparison of different planning methods. 
The preference-agnostic trajectories for the RLC case in Fig.~\ref{fig:long-term_comparison}(a) and (b) conflict with user-preferred clearance.
Specifically, in Fig.~\ref{fig:long-term_comparison}(b), for $\mathcal{E_M} = \text{RLC}$, the planned trajectory passes in front of RLLC. 
In contrast, as shown in Fig.~\ref{fig:long-term_comparison}(c), this trajectory conflicts with the user-preferred clearances. 
Accordingly, as shown in Fig.~\ref{fig:long-term_comparison}(d), the trajectory becomes feasible when it is planned to pass behind the RLLC.
Finally, the method of \cite{werling2010optimal} fails to find a feasible solution due to its conservative behavior.

For further comparison, we conduct a quantitative evaluation of how well our system aligns with user-preferred clearances.
We consider two scenarios on a two-lane road, where all vehicles travel at 45 kph and are randomly assigned to a lane.
In Scenario 1, vehicles keep a 15 m gap to reflect the user-preferred clearances of discouraging overtaking, whereas in Scenario 2, vehicles maintain a 25 m gap, providing sufficient space for EC to overtake.
For each scenario, we run 30 closed-loop simulations in HighwayEnv \cite{leurent2018environment}, each lasting 30 seconds.
We use \eqref{equ:cost} to select or optimize trajectories for all methods.
Table \ref{tab:long-term_comparison} summarizes the results.

In Scenario 1, our planner stays in its lane due to limited space for overtaking under the user-preferred clearances, as shown in Fig.~\ref{fig:scenario_result}; thus, the lane change count is close to zero.
In contrast, in Scenario 2, sufficient space is available; thus, our planner overtakes while still reflecting the user-preferred clearances.
Other planners tend to overtake in both scenarios, as they do not account for preferred clearances.
The Frenet-planner \cite{werling2010optimal} fails to find the collision-free trajectory in scenario 1.
The Batch MPC \cite{adajania2022multi} and CBF + MPPI \cite{jin2023safety} find collision-free trajectories but do not consider the user-preferred clearances; thus, they exhibit high lane change counts and speeds in both scenarios.
Consequently, only our planner satisfies the preferred clearances in both scenarios.
In Scenario 2, the slight 0.14$\%$ reduction in our planner's preference rate is attributed to the inclusion of a slack variable in $\mathcal{S}_\phi$ as defined in \eqref{equ:app_preference_opt}.

\begin{figure}[!t]
\captionsetup{singlelinecheck = false, font=small, labelfont=bf, skip=-2pt}
\captionsetup[subfloat]{singlelinecheck = true, labelfont=scriptsize,textfont=scriptsize}
\centering
\vspace{5pt}
\includegraphics[height=3.5cm]{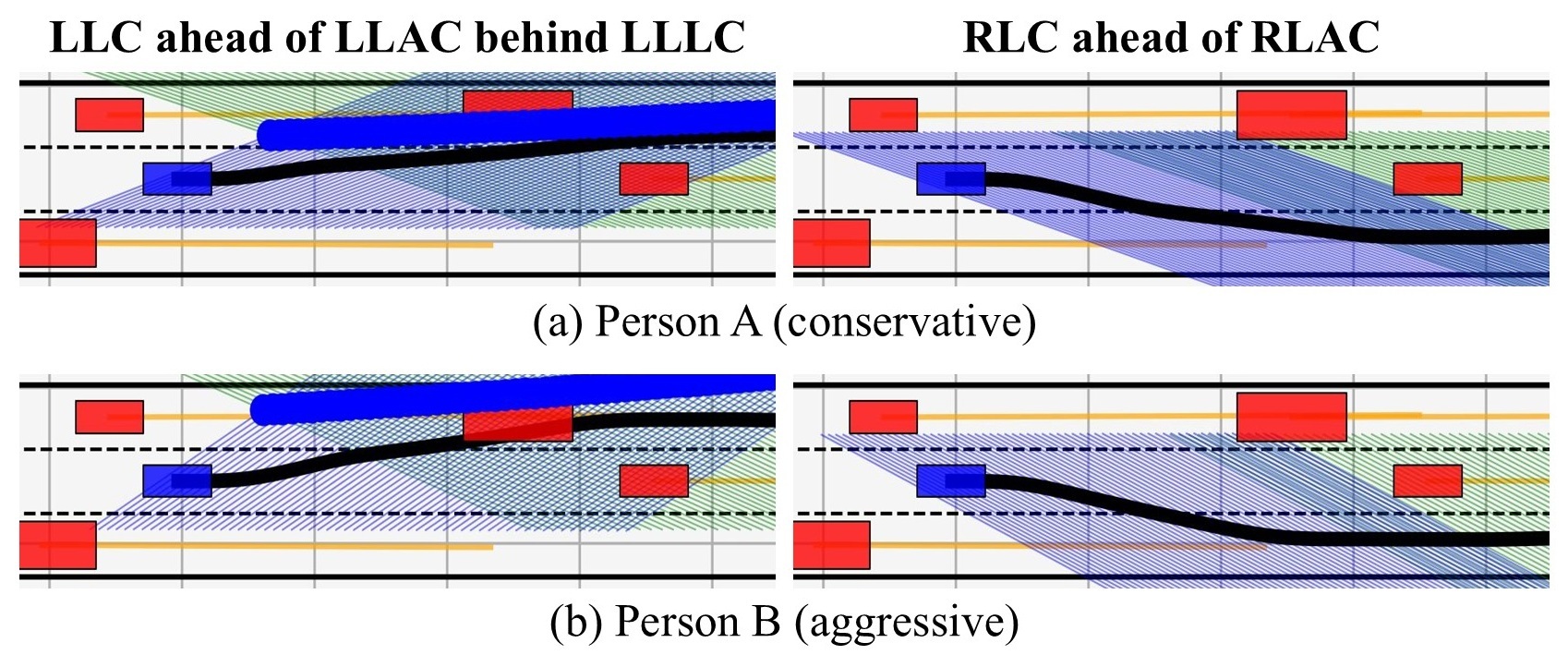}
\caption{
The {red} and {blue} boxes, the black point, the {orange} and {black} solid lines, and the light blue and green lines are the same as in Fig. \ref{fig:long-term_comparison}.
(a) and (b) show the trajectory candidates in the same driving scene. 
Person A identified himself as a conservative driver in the questionnaire, and his preference space is constructed as shown in (a). 
Similarly, Person B identified himself as an aggressive driver, and his preference space is constructed as shown in (b).
The results indicate that as the driver profile becomes more aggressive, $\mathcal{S}_\phi$ enlarges, leading to an optimized trajectory that behaves more assertively.
}
\label{fig:scenario_result2}
\vspace{-10pt}
\end{figure}

\begin{table}[!t]
\vspace{10pt}
\centering
\caption{Results of driving 30 closed-loop simulations.}
\small
\setlength{\tabcolsep}{8pt}  
\setlength{\extrarowheight}{0pt}
\renewcommand{\arraystretch}{1.2}  
\begin{tabular}{l | c c c}
    \midrule
    \multirow{2}{*}{\textbf{Driving style}}  
        & {\textbf{Pref.}} & {\textbf{Avg.}} & {\textbf{Change}} \\
        & {\textbf{rate [\%]}} & {\textbf{speed [m/s]}} & {\textbf{count}} \\
    \midrule
    Conservative         & 99.98  & 13.90 & 1.33  \\
    Aggressive         & 100  & 14.23  & 1.87 \\
    \midrule
\end{tabular}
\label{tab:ablataion}

\vspace{0pt}
\begin{flushleft}
\justifying
\footnotesize 
\textbf{Pref. rate}, \textbf{Avg. speed}, and \textbf{Change count} are the same as in Table~\ref{tab:long-term_comparison}.
\end{flushleft}
\vspace{-10pt}
\end{table}

\subsection{Evaluation on different driving types}
To validate whether personalized driving can be achieved by incorporating responses to the questionnaire from different users, we consider two driver types, conservative and aggressive. 
We set the responses so that the conservative driver maintains a larger preferred clearance for safety compared to the aggressive driver.
As shown in Fig.~\ref{fig:scenario_result2}, conservative and aggressive drivers exhibit different $\mathcal{S}_\phi$ for the same driving scene.
Consequently, the optimized trajectory differs because the aggressive driver, having a larger $\mathcal{S}_\phi$, attempts lane changes more assertively than the conservative driver.

For quantitative analysis, we use the responses of the conservative and aggressive drivers to run 30 closed-loop simulations in HighwayEnv. 
Each simulation lasts 30 seconds in a three-lane environment with big and small vehicles randomly placed, driving at 45~kph.
Table~\ref{tab:ablataion} summarizes the results.
Both driving results accurately reflect the user-preferred clearance. 
The results using the aggressive driver response show a higher average speed and lane change count compared to those using the conservative driver response, because $\mathcal{S}_\phi$ constructed from the aggressive driver's response is larger.
The slight 0.02$\%$ reduction in the preference rate for the conservative driver is attributed to the inclusion of a slack variable in $\mathcal{S}_\phi$ as defined in \eqref{equ:app_preference_opt}.

%% file: Conclusion.tex
\section{Conclusion}
\label{sec:conclusion}
Ignoring user preference while driving leads to discomfort and reduces the trust in ADS for the user.
In this study, we propose the preference-aware planning framework based on the user-preferred clearance in road driving. 
We design the questionnaire to obtain the user-preferred clearances. 
Furthermore, we construct the user preference space from the questionnaire responses and perform trajectory planning within this space to generate plans that reflect the user-preferred clearances.
This allows the evaluation of performance by measuring the proportion of total driving time that the vehicle spends within the preference space derived from the user responses.
We validate our approach through simulation by comparing it against various planners, focusing on their ability to account for user preferences.
Future work will focus on extending the framework beyond user preferences in the vehicle interaction to incorporate context-aware preferences (e.g., I'm late! I'm in a hurry, so go fast)  while guaranteeing preference satisfaction and ensuring collision avoidance.